\documentclass[10pt,twocolumn,letterpaper]{article}

\usepackage[pagenumbers]{cvpr} 

\usepackage{multirow}
\usepackage{bm}

\definecolor{White}{rgb}{1.,0.,1.}
\definecolor{first}{rgb}{.8,.0,.0}
\definecolor{second}{rgb}{.0,.6,.0}
\definecolor{third}{rgb}{.0,.0,.8}

\definecolor{ceiling}{RGB}{214,  38, 40}
\definecolor{floor}{RGB}{43, 160, 4}
\definecolor{wall}{RGB}{158, 216, 229}
\definecolor{window}{RGB}{114, 158, 206}
\definecolor{chair}{RGB}{204, 204, 91}
\definecolor{bed}{RGB}{255, 186, 119}
\definecolor{sofa}{RGB}{147, 102, 188}
\definecolor{table}{RGB}{30, 119, 181}
\definecolor{tvs}{RGB}{160, 188, 33}
\definecolor{furniture}{RGB}{255, 127, 12}
\definecolor{objects}{RGB}{196, 175, 214}

\definecolor{car}{rgb}{0.39215686, 0.58823529, 0.96078431}
\definecolor{bicycle}{rgb}{0.39215686, 0.90196078, 0.96078431}
\definecolor{motorcycle}{rgb}{0.11764706, 0.23529412, 0.58823529}
\definecolor{truck}{rgb}{0.31372549, 0.11764706, 0.70588235}
\definecolor{othervehicle}{rgb}{0.39215686, 0.31372549, 0.98039216}
\definecolor{person}{rgb}{1.        , 0.11764706, 0.11764706}
\definecolor{bicyclist}{rgb}{1.        , 0.15686275, 0.78431373}
\definecolor{motorcyclist}{rgb}{0.58823529, 0.11764706, 0.35294118}
\definecolor{road}{rgb}{1.        , 0.        , 1.        }
\definecolor{parking}{rgb}{1.        , 0.58823529, 1.        }
\definecolor{sidewalk}{rgb}{0.29411765, 0.        , 0.29411765}
\definecolor{otherground}{rgb}{0.68627451, 0.        , 0.29411765}
\definecolor{building}{rgb}{1.        , 0.78431373, 0.        }
\definecolor{fence}{rgb}{1.        , 0.47058824, 0.19607843}
\definecolor{vegetation}{rgb}{0.        , 0.68627451, 0.        }
\definecolor{trunk}{rgb}{0.52941176, 0.23529412, 0.        }
\definecolor{terrain}{rgb}{0.58823529, 0.94117647, 0.31372549}
\definecolor{pole}{rgb}{1.        , 0.94117647, 0.58823529}
\definecolor{trafficsign}{rgb}{1.        , 0.        , 0.        }
\definecolor{otherstructure}{rgb}{0.98039215, 0.58823529, 0.}
\definecolor{otherobject}{rgb}{0.19607843, 1.        , 1.        }

\makeatletter
\newcommand{\car@semkitfreq}{3.92}
\newcommand{\bicycle@semkitfreq}{0.03}
\newcommand{\motorcycle@semkitfreq}{0.03}
\newcommand{\truck@semkitfreq}{0.16}
\newcommand{\othervehicle@semkitfreq}{0.20}
\newcommand{\person@semkitfreq}{0.07}
\newcommand{\bicyclist@semkitfreq}{0.07}
\newcommand{\motorcyclist@semkitfreq}{0.05}
\newcommand{\road@semkitfreq}{15.30}
\newcommand{\parking@semkitfreq}{1.12}
\newcommand{\sidewalk@semkitfreq}{11.13}
\newcommand{\otherground@semkitfreq}{0.56}
\newcommand{\building@semkitfreq}{14.1}
\newcommand{\fence@semkitfreq}{3.90}
\newcommand{\vegetation@semkitfreq}{39.3}
\newcommand{\trunk@semkitfreq}{0.51}
\newcommand{\terrain@semkitfreq}{9.17}
\newcommand{\pole@semkitfreq}{0.29}
\newcommand{\trafficsign@semkitfreq}{0.08}
\newcommand{\semkitfreq}[1]{{\csname #1@semkitfreq\endcsname}}

\definecolor{cvprblue}{rgb}{0.21,0.49,0.74}
\usepackage[pagebackref,breaklinks,colorlinks,allcolors=cvprblue]{hyperref}

\def\paperID{***} 
\def\confName{CVPR}
\def\confYear{2025}

\title{SSEditor: Controllable Mask-to-Scene Generation with Diffusion Model}

\author{Haowen Zheng, YanyanLiang\\
Macau University of Science and Technology\\
{\tt\small zhengnayin@gmail.com, yyliang@must.edu.mo}
}

\begin{document}
\twocolumn[{
\renewcommand\twocolumn[1][]{#1}
\maketitle
\begin{center}
    \captionsetup{type=figure}
    \includegraphics[width=\textwidth]{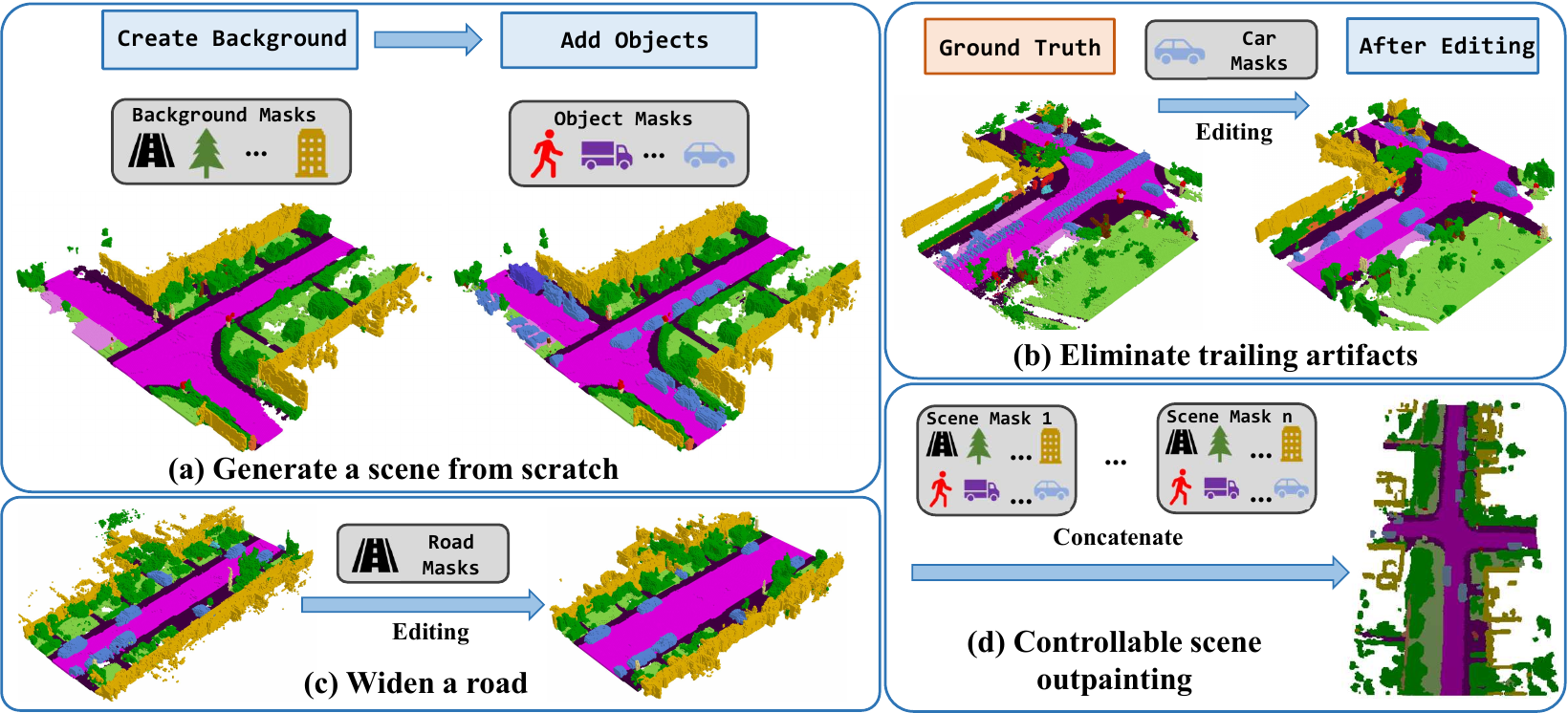}
    \captionof{figure}{Controllable 3D semantic scene generation by SSEditor. The proposed SSEditor enables users to customize the generation or editing of 3D scenes using pre-built mask assets: (a) create a background scene and generate objects on it; (b) eliminate trailing artifacts of dynamic objects in SemanticKITTI \cite{behley2019semantickitti}; (c) modify roads, such as expanding a two-lane road to a four-lane road; (d) concatenate masks from various scenes to produce a larger-scale 3D scene.
    }
    \label{fig1_overview}
\end{center}
}]

\begin{abstract}
Recent advancements in 3D diffusion-based semantic scene generation have gained attention. However, existing methods rely on unconditional generation and require multiple resampling steps when editing scenes, which significantly limits their controllability and flexibility. To this end, we propose \textbf{SSEditor}, a controllable \textbf{S}emantic \textbf{S}cene \textbf{Editor} that can generate specified target categories without multiple-step resampling. SSEditor employs a two-stage diffusion-based framework: (1) a 3D scene autoencoder is trained to obtain latent triplane features, and (2) a mask-conditional diffusion model is trained for customizable 3D semantic scene generation. In the second stage, we introduce a geometric-semantic fusion module that enhance the model's ability to learn geometric and semantic information. This ensures that objects are generated with correct positions, sizes, and categories. Extensive experiments on SemanticKITTI and CarlaSC demonstrate that SSEditor outperforms previous approaches in terms of controllability and flexibility in target generation, as well as the quality of semantic scene generation and reconstruction. More importantly, experiments on the unseen Occ-3D Waymo dataset show that SSEditor is capable of generating novel urban scenes, enabling the rapid construction of 3D scenes.
\end{abstract}
    
\section{Introduction}
\label{sec:intro}

In recent years, 3D diffusion models have made notable achievements in generating both indoor \cite{tang2024diffuscene, ju2024diffindscene, zhai2024commonscenes} and outdoor \cite{liu2023pyramid, ren2024xcube, wang2024occsora, lee2023diffusion, lee2024semcity} environments, as well as a single object \cite{zhou20213d, shue20233d, karnewar2023holodiffusion}. Compared to indoor scenes and individual objects, outdoor scenes present more challenges due to their sparser and more complex representations. For instance, voxel-based representations of outdoor environments often contain a significant number of empty voxels. Moreover, outdoor environments contain smaller targets, such as pedestrians and cyclists, further complicating the generation process. While voxel-based representations \cite{liu2023pyramid, ren2024xcube, wang2024occsora, lee2023diffusion} provide a straightforward approach to modeling 3D semantic scenes, they suffer from redundancy in empty regions and high computational cost. To mitigate these issues, the triplane representation \cite{chan2022efficient} is utilized to reduce unnecessary information in 3D outdoor scenes \cite{lee2024semcity}. Although these methods have shown promising results, they still face several limitations.

The primary limitation lies in their weak controllability. Unconditional generation restricts the ability to guide the creation of 3D scenes, while conditioning on the entire scene (e.g., scene refinement based on ground truth) is overly rigid. This lack of flexible control leads to another drawback: editing specific local regions, such as adding or removing objects, necessitates masking non-target areas and employing a multi-step resampling process for repainting \cite{lugmayr2022repaint}. It significantly increases generation time. Despite the use of this resampling strategy, repainting remains uncontrollable and often fails to produce the desired results.

To address the aforementioned challenges, we propose SSEditor, a flexible and controllable two-stage framework for semantic scene generation based on the latent diffusion model (LDM) \cite{rombach2022high}. In the first stage, we train a 3D scene autoencoder to learn triplane features via semantic scene reconstruction. In the second stage, we train a mask conditional diffusion model on the triplane features. Specifically, to enable the customizable generation of 3D semantic scenes, we present a Geometric-Semantic Fusion Module (GSFM), which consists of a geometric branch and a semantic branch. The geometric branch encodes 3D masks that represent an object's position, size, and orientation, while the semantic branch processes semantic labels and tokens for providing coarse and fine-grained semantic information. The semantic tokens are generated from the features of a specific category. These features are then aggregated and integrated into the cross-attention module of the diffusion model, enhancing its perception of both geometric and semantic information. Benefiting from the above design, SSEditor effectively accomplishes the mask-to-semantic scene generation task.

In addition, we create a 3D mask asset library encompassing various categories to facilitate custom scene generation during inference. The 3D masks in the library are stored in the form of trimasks, which are composed of three orthogonal 2D planes derived from the decomposition of the 3D mask. As shown in Fig. \ref{fig1_overview}, users can choose from a range of assets, such as cross-shaped roads, vehicles, pedestrians, and cyclists, to generate their desired 3D semantic scenes. The assets can also be edited to simulate more urban scenarios, such as expanding a two-lane road to four or more lanes.

Our contributions can be summarized into three points:

\begin{itemize}
\item We propose SSEditor, a controllable mask-to-scene generation framework that enables users to easily customize and generate 3D semantic scenes using various assets.

\item We propose GSFM to integrate geometric and semantic information. In GSFM, the geometric branch encodes 3D masks as embeddings to accurately control the position, size, and orientation of objects, while the semantic branch processes semantic labels and tokens for improved class control of the generated targets.

\item Experiments on outdoor datasets demonstrate that our proposed method achieves superior generation quality and reconstruction performance. Furthermore, qualitative results indicate that SSEditor can controllably perform various downstream tasks, such as scene inpainting, resource expansion, novel urban scene generation, and removal of trailing artifacts.
\end{itemize}

\section{Related Work}
\label{sec:related_work}

\begin{figure*}[t]
\begin{center}
\includegraphics[width=\linewidth]{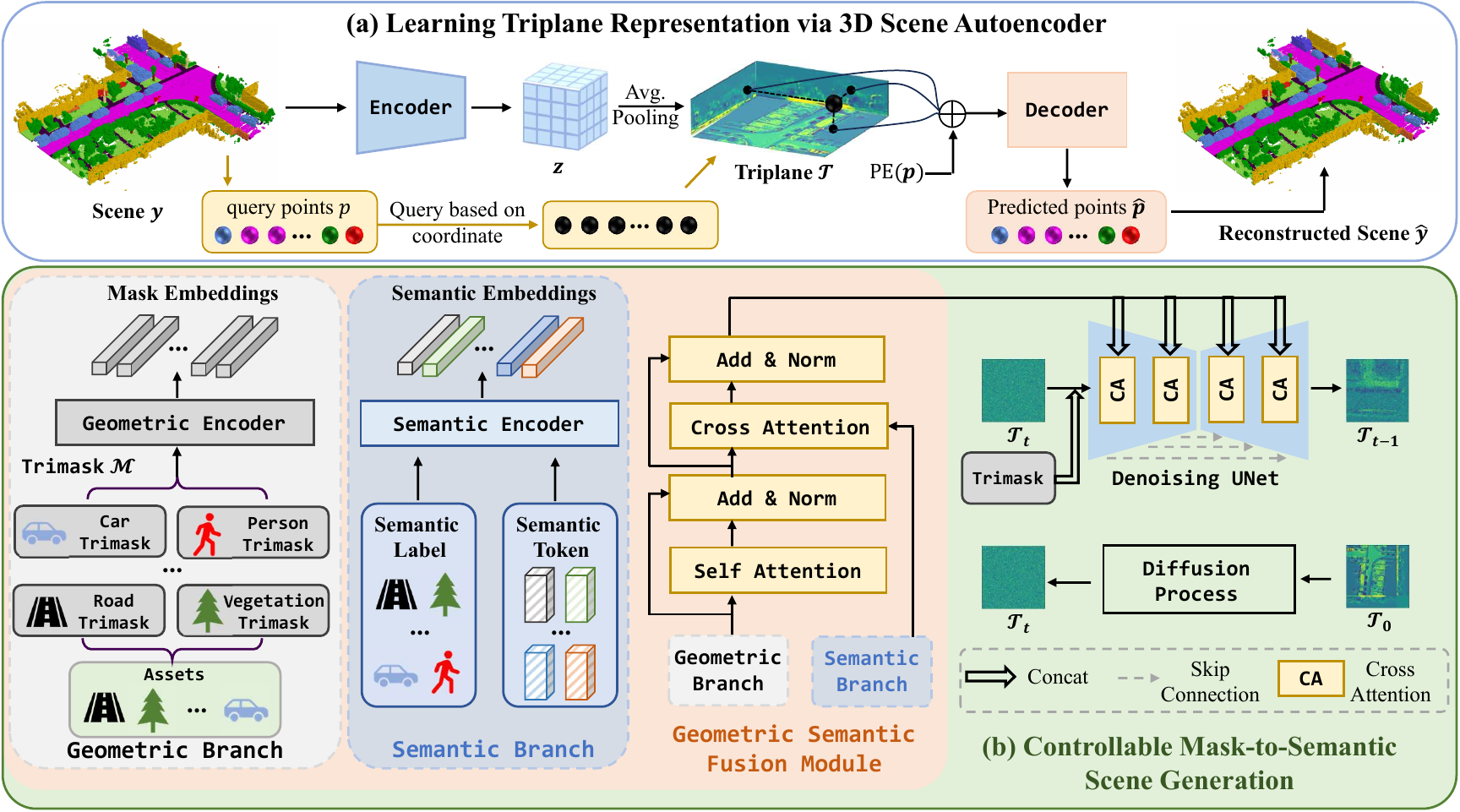}
\end{center}
\caption{Illustration of our SSEditor framework. It comprises two main processes: (a) a 3D autoencoder learns the triplane representation via scene reconstruction, and (b) controllable semantic scene generation is achieved through masks, semantic labels, and tokens. The Geometric-Semantic Fusion Module is essential for the diffusion model to effectively learn both geometric and semantic information.}
\label{fig:framework}
\end{figure*}

\textbf{Controllable Diffusion Models.} Denoising diffusion probabilistic models (DDPM) \cite{ho2020denoising} inspires a series of diffusion-based controllable generation approaches. Text-guided image generation shows strong capabilities in image editing tasks, such as inpainting \cite{avrahami2022blended, nichol2021glide, ramesh2022hierarchical} and outpainting \cite{saharia2022palette}. In addition, several studies incorporate more control signals, such as layouts \cite{zheng2023layoutdiffusion}, semantic maps \cite{wang2022pretraining, gafni2022make, rombach2022high, zhang2023adding}, to facilitate image generation. Building on these advancements, controllable diffusion models have been further extended to the 3D domain. These models can leverage images \cite{choy20163d, xu2019disn}, text \cite{li2023diffusion, michel2022text2mesh}, partial point clouds \cite{nakayama2023difffacto} or multi-modal conditions (e.g., text-image or text-voxels) \cite{mittal2022autosdf, vahdat2022lion} to guide the generation of a single 3D object. However, the aforementioned controllable generative models can only be applied to 2D images or individual 3D objects, making it challenging for them to handle complex large-scale 3D scenes.

\noindent\textbf{3D Semantic Scene Generation.} 3D semantic scene generation can be categorized into indoor and outdoor scene generation. CommonScenes \cite{zhai2024commonscenes} generates indoor scenes based on scene graphs. DiffuScene \cite{tang2024diffuscene} performs indoor scene generation and completion based on a text prompt or incomplete 3D targets. InstructScene \cite{lin2024instructscene} incorporates user instructions into semantic graph priors and decodes them into 3D indoor scenes. Build-A-Scene \cite{eldesokey2024build} enables users to flexibly create indoor scenes by adjusting layouts. In contrast, outdoor scene generation is more complex, which features diverse objects, more occlusions, and varying distances. \cite{lee2023diffusion} generates 3D multi-object scenes in simulated outdoor environments, while PDD \cite{liu2023pyramid} employs a coarse-to-fine strategy to further improve generation quality. For more complex real-world outdoor scenes, SemCity \cite{lee2024semcity} uses triplane diffusion to achieve unconditional generation or conditional 3D occupancy refinement.

Due to the significant differences between indoor and outdoor environments, these controllable indoor scene generation methods \cite{tang2024diffuscene, lin2024instructscene, eldesokey2024build} are difficult to apply to outdoor scenes. For outdoor environments, \cite{liu2023pyramid, lee2024semcity} can only refine scenes by conditionally inputting the entire 3D layout. Moreover, when conducting scene inpainting, SemCity \cite{lee2024semcity} requires multiple-step resampling \cite{lugmayr2022repaint} and lacks one-step sampling capability. Additionally, it can not control the categories of the generated regions. This lack of flexible control prevents users from generating their desired scenes. In this paper, our proposed SSEditor overcomes these limitations and enables users to generate large-scale outdoor scenes from masks with traditional DDPM sampling \cite{ho2020denoising}.

\section{Method}

In this paper, we propose our SSEditor, as illustrated in Fig. \ref{fig:framework}. The primary objective of SSEditor is to enable users to generate 3D outdoor semantic scenes with flexibility and controllability. To achieve this goal, we first leverage a 3D scene autoencoder to learn the triplane representation (Sec. \ref{sec:3.1}) and then create an asset library for storing 3D masks (Sec. \ref{sec:3.2}). To enhance the accuracy for generating the positions, sizes, and categories of target objects, we implement a geometry-semantic fusion module that improves the model's understanding of geometric and semantic information, facilitating our controllable mask-to-scene generation. (Sec. \ref{sec:3.3}). During inference, users can flexibly select or create assets to customize 3D scene construction, such as controllable inpainting, novel urban scene generation and trailing artifacts removal (Sec. \ref{sec:3.4}).

\subsection{3D Scene Autoencoder with Triplane}
\label{sec:3.1}
Fig. \ref{fig:framework}(a) illustrates that the 3D scene autoencoder learns the triplane representation through scene reconstruction. We employ an encoder composed of 3D convolutions to encode a given scene $\textbf{y} \in \mathbb{R}^{X \times Y \times Z}$ into $\textbf{z} \in \mathbb{R}^{C_z \times \frac{X}{d} \times \frac{Y}{d} \times \frac{Z}{d_z}}$, where $C_z$, $X$, $Y$ and $Z$ denote the number of channel and the resolution of 3D voxel space, while $d$ and $d_z$ indicate the down-sampling factors. Axis-wise average pooling is then applied across the three dimensions of $\textbf{z}$ to derive the triplane representation $\mathcal T = [\mathcal T^{xy}, \mathcal T^{xz}, \mathcal T^{yz}]$. In addition, we sample query points $\textbf{p}$ from the scene voxels and aggregate the corresponding triplane features based on their coordinates, which can be represented as $\mathcal T(\textbf{p})=\mathcal T^{xy}(\textbf{p}^{xy})+\mathcal T^{xz}(\textbf{p}^{xz})+\mathcal T^{yz}(\textbf{p}^{yz})$. The aggregated triplane features, combined with positional embedding, are decoded to obtain the predicted points $\bm{\hat p}$. The predicted points reconstruct the scene $\bm{\hat y}$ based on the original coordinate information.
The autoencoder is trained with a joint loss $\mathcal{L}_{AE}$, including the cross-entropy loss $\mathcal{L}_{CE}$ \cite{roldao2020lmscnet} on the points, and the Lov\'{a}sz-softmax loss $\mathcal{L}_{Lov}$ \cite{berman2018lovasz} on the reconstructed scene:

\begin{equation}
    \mathcal{L}_{AE} = \mathcal{L}_{CE}(\bm{\hat p}, \textbf{p}) +  \alpha \mathcal{L}_{Lov}(\bm{\hat y}, \textbf{y})
\label{eq1}
\end{equation}
where $\alpha$ is a loss weight.

\begin{figure}[t]
\begin{center}
\includegraphics[width=\linewidth]{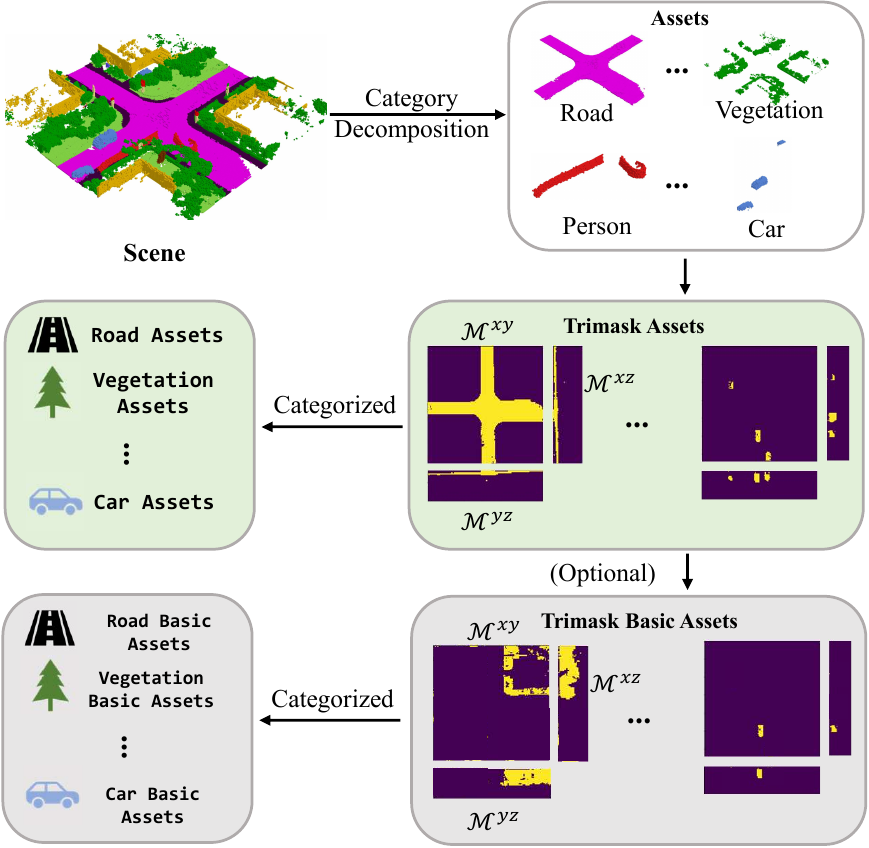}
\end{center}
\caption{Pipeline of building 3D mask assets. The 3D mask is stored in the corresponding category in the form of a trimask.}
\label{fig:assets}
\end{figure}

\subsection{3D Mask Assets}
\label{sec:3.2}
To achieve a customizable generation of 3D scenes, controlling conditions need to be user-friendly inputs that can accurately reflect information such as target position and size. A 3D mask effectively serves this purpose. By utilizing the triplane representation, as illustrated in Fig. \ref{fig:assets}, we compress the 3D voxel mask into three 2D orthogonal planes, forming a trimask. The trimask can be represented as $\mathcal{M}=[\mathcal{M}^{xy}, \mathcal{M}^{xz}, \mathcal{M}^{yz}]$. All categories in the scene are decomposed into trimasks and stored in corresponding asset libraries. In addition to these scene-level assets, we also provide a basic version of the assets, which contains individual or segmented assets. This allows users to more conveniently utilize the basic assets to customize and construct scene-level assets. More importantly, users can also draw masks directly within the basic assets or scene-level assets. For example, the assets collected in the dataset only include small roads (2-lane and 4-lane). Users can edit the basic road assets (e.g., by copying, translating, or rotating) to create wider lanes, such as 6-lane or 8-lane roads, to support the generation of more complex 3D scenes.

\subsection{Controllable Mask-to-Scene Generation}
\label{sec:3.3}
The trimasks in the established assets offer valuable geometric information, including position, orientation, and scale. However, this is not enough for effective mask-to-semantic scene generation. We also need to extract detailed semantic information to ensure accurate object category generation. To tackle this, we propose a Geometric-Semantic Fusion Module (GSFM), as shown in Fig. \ref{fig:framework}(b), which consists of two branches: a geometric branch and a semantic branch.

\noindent \textbf{Geometric Branch.} The geometric branch encodes the trimask into mask embedding using an multi-layer perception (MLP), consisting of two linear layers and one activation layer. For simplicity, we first concatenate the trimask into a 2D feature maps $\mathcal{M'} \in \mathbb{R}^{N \times ({X_m}+{Z_m}) \times ({Y_m}+{Z_m})}$, where N is the number of semantic classes, ${X_m}=\frac{X}{d}$, ${Y_m}=\frac{Y}{d}$ and ${Z_m}=\frac{Z}{d_z}$. The mask embedding $E_m \in \mathbb{R}^{N \times C_{emb}}$ can be obtained by

\begin{align}
& \text{MLP}(x) = \text{Linear}(\text{GeLU}(\text{Linear}(x) \\
& E_m = \text{MLP}(\mathcal{M'})
\end{align}

The extracted mask embeddings currently operate independently and lack geometric information from other categories. To resolve this, we employ self-attention to capture the geometric relationships between masks of different categories through Eq. \ref{eq3}. This allows the model to detect targets of varying scales within the same category and identify overlapping regions between different category masks.

\begin{equation}
    E'_m = E_m + \text{LayerNorm}(\text{SelfAttn}(E_m)).
\label{eq3}
\end{equation}

\noindent \textbf{Semantic Branch.} In the semantic branch, we begin with an embedding layer to convert semantic labels into label embeddings $E_{label} \in \mathbb{R}^{N \times C_{emb}}$. However, the label embeddings offer only coarse-grained semantic information, which is inadequate for large-scale scene generation. The voxels generated within the mask regions may still contain a number of incorrect categories. To address this, we introduce a finer-grained semantic token $\textbf{T}_{sem} \in \mathbb{R}^{N \times C_{emb}}$, which is defined as:

\begin{equation}
    \textbf{T}_{sem}^i = \text{Spatial Pooling}(\mathcal{M}_i \cdot \mathcal{T})
\end{equation}
where $i$ indicates the $i$-th semantic class and spatial pooling represents average pooling along the spatial dimension. As a result, the semantic embeddings $E_{sem} \in \mathbb{R}^{N \times C_{emb}}$ can be formulated as 
\begin{equation}
    E_{sem} = \text{MLP} (E_{label} + \textbf{T}_{sem})
\end{equation}

\noindent \textbf{Geometric-Semantic Fusion Module.} 
The GSFM integrates mask embeddings $E_m$ and semantic embeddings $E_{sem}$ through cross-attention. We use the geometric embeddings as the query $Q \in \mathbb{R}^{N \times C_{emb}}$ and concatenate the geometric and semantic embeddings to form the key $K$ and value $V \in \mathbb{R}^{2N \times C_{emb}}$. The fused embeddings $E_{fused}$ can then be represented as:

\begin{equation}
    E_{fused} = E'_m + \text{LayerNorm}(\text{CrossAttn}(Q, K, V))
\label{eq7}
\end{equation}

\noindent \textbf{Mask Conditional Diffusion Model.} Following LDM \cite{rombach2022high}, we conduct diffusion and denoising process on the triplane features $\mathcal{T}$ to learn our mask conditional diffusion model $D_\theta$. We add t steps of Gaussian noise to a clean triplane features $\mathcal{T}_0$ and obtain a noised triplane $\mathcal{T}_t \sim q(\mathcal{T}_t|\mathcal{T}_0)=\mathcal{N}(\sqrt{\overline{\alpha}_t}\mathcal{T},(1-\overline{\alpha}_t)\textbf{I})$, where $\mathcal{N}$ is the Gaussian distribution, $\overline{\alpha}_t=\prod_{i=1}^t \alpha_i$ and $\alpha_t=1-\beta_t$ with a variance schedule $\beta_t$. Then the diffusion model $D_\theta$ can be trained with the mean-squared error loss:

\begin{equation}
    \mathcal{L} = \mathbb{E}_{t\sim[1,T]}\Vert\mathcal{T}_0-D_\theta(\text{Concat}(\mathcal{T}_t, \mathcal{M}),t)\Vert_2
\end{equation}

To support mask conditional generation, we inject the fused embedding $E_{fused}$ obtained from Eq. \ref{eq7} into the cross attention of the diffusion model. In addition, we concatenate the trimask with $\mathcal{T}_t$ to further enhance the guidance of the mask. Following classifier-free guidance \cite{ho2022classifier}, we randomly set the trimask to zero during training to simulate the effect of not using the trimask.

\begin{table}[t]
\centering
\resizebox{\linewidth}{!}{
\begin{tabular}{l|ccccc}
\toprule[1pt]
Model & FID $\downarrow$ & KID $\downarrow$ & IS $\uparrow$ & Prec. $\uparrow$ & Rec. $\uparrow$ \\ \midrule
\multicolumn{6}{c}{SemanticKITTI \cite{behley2019semantickitti}}      \\ \midrule
SSD \cite{lee2023diffusion}  & 117.46 & 0.12 & 2.15 $\pm$ 0.13 & 0.01 & 0.08 \\
SemCity \cite{lee2024semcity}  & 61.20 & 0.04 & 2.43 $\pm$ 0.11 & 0.19 & 0.12 \\
SSEditor (ours)  & \textbf{47.93} & \textbf{0.03} & \textbf{2.55 $\pm$ 0.14} & \textbf{0.31} & \textbf{0.51} \\ \midrule
\multicolumn{6}{c}{CarlaSC \cite{wilson2022motionsc}}      \\ \midrule
SSD \cite{lee2023diffusion}  & 148.14 & 0.18 & 2.23 $\pm$ 0.10 & 0.15 & 0.01 \\
SemCity \cite{lee2024semcity}  & 137.94 & 0.13 & \textbf{3.03 $\pm$ 0.17} & 0.20 & 0.02 \\
SSEditor (ours)  & \textbf{50.98} & \textbf{0.03} & 2.28 $\pm$ 0.08 & \textbf{0.37} & \textbf{0.18} \\
\bottomrule[1pt]
\end{tabular}
}
\caption{Quantitative results on SemanticKITTI and CarlaSC. The metrics are measured between the rendered image of the  generated scene and the real scene. Prec. and Rec. indicates precision and recall, respectively.}
\label{tab1}
\end{table}

\subsection{Downstream Applications}
\label{sec:3.4}
Unlike unconditional scene generation \cite{lee2024semcity}, our SSEditor can flexibly handle various downstream tasks based on the created assets, such as controllable scene inpainting and controllable scene outpainting. Note that our method does not require a resampling strategy \cite{lugmayr2022repaint}.

\noindent\textbf{Controllable Scene Inpainting} can facilitate basic scene editing, such as adding or removing objects. Based on this, SSEditor can simulate corner cases in autonomous driving scenarios, such as vehicle congestion at intersections, bicycles haphazardly parked on the roadside, and pedestrians crossing the street. Furthermore, the accumulation of multiple LiDAR frames causes trailing artifacts in dynamic objects within the SemanticKITTI dataset \cite{behley2019semantickitti}. Our SSEditor effectively resolves this issue. In addition, by editing background assets such as roads and sidewalks, SSEditor can also widen roads to simulate scenarios with greater traffic. 

\noindent\textbf{Controllable Scene Outpainting} can assist in scene extension. By selecting appropriate background assets and combining them, such as stitching together continuous roads, we can controllably extend the scene.

\noindent\textbf{Novel Urban Scene Generation} enables the rapid construction of 3D occupancy datasets. Imagine that we want to build a 3D semantic scene for a new city: we can create different assets based on LiDAR point clouds, and then generate a novel urban scene based on these assets.

\noindent\textbf{Removing trailing artifacts.} SemanticKITTI \cite{behley2019semantickitti} aggregates multiple LiDAR frames to create dense 3D occupancy scenes, but this introduces trailing artifacts for moving objects in the ground truth, as shown in Fig. \ref{fig1_overview}(b). Our method can effectively remove these artifacts and utilizes existing object assets to generate new objects.

\begin{table}[t]
\centering
\resizebox{\linewidth}{!}{
\begin{tabular}{l|c|cc}
\toprule[1pt]
Model & Input & IoU $\uparrow$ & mIoU $\uparrow$  \\ \midrule
Symphonies \cite{jiang2024symphonize}  & RGB & 41.92 & 14.89 \\
SCPNet \cite{xia2023scpnet}  & Point Cloud & 50.24 & 37.55  \\
SSEditor (ours)  & 3D Mask & \textbf{57.85} & \textbf{43.09}  \\ 
\bottomrule[1pt]
\end{tabular}
}
\caption{Quantitative results on SemanticKITTI validation set. IoU and mIoU indicate how effectively our method handles geometric information and comprehends semantic information during generation, respectively.}
\label{tab2}
\end{table}
\section{Experiments}

\begin{figure}[t]
\begin{center}
\includegraphics[width=\linewidth]{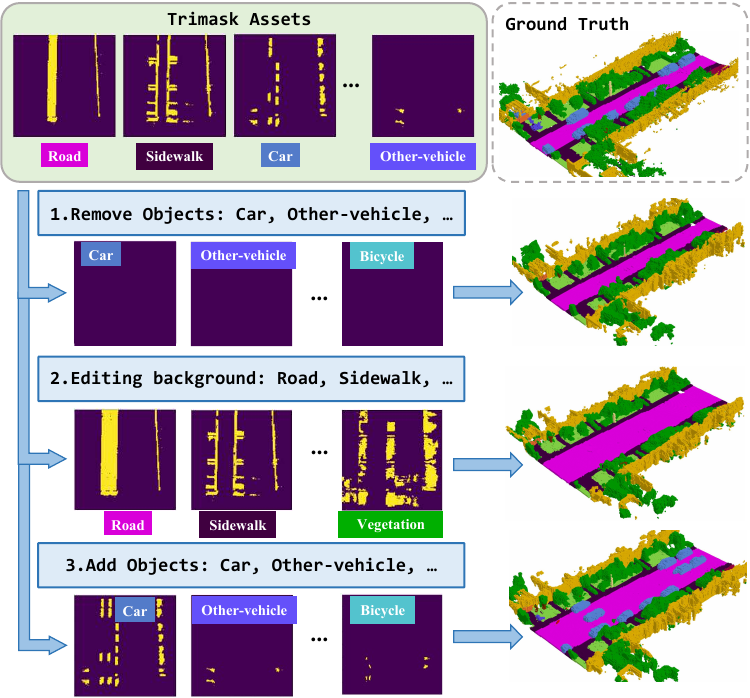}
\end{center}
\caption{The details of editing 3D scenes with SSEditor: 1. When the mask of an object is set to 0, the corresponding object can be completely removed. 2. The background can be edited, such as widening roads to simulate heavier traffic. 3. Objects can be added to the edited scene.}
\label{fig:edit_detail}
\end{figure}

\subsection{Datasets}
We conduct our experiments on the SemanticKITTI \cite{behley2019semantickitti} and CarlaSC \cite{wilson2022motionsc} datasets. SemanticKITTI dataset is a large-scale real-world benchmark for semantic scene understanding in autonomous driving. It contains 20 semantic classes. Each scene is represented by a 256$\times$256$\times$32 voxel grid with a voxel resolution of 0.2m. CarlaSC dataset is a synthetic dataset with labels for 11 semantic classes, generated using the CARLA simulator. Each scene has a resolution of 128$\times$128$\times$8, covering an area of 25.6 meters around the vehicle, with a height of 3 meters. Additionally, we validated the cross-dataset transferability of SSEditor on Occ3D-Waymo \cite{tian2024occ3d}. We only included the occupancy labels from Occ3D-Waymo \cite{tian2024occ3d} as trimasks in our asset library and then simulated the generation of unknown urban scenes. Note that we disregard the Occ3D-Waymo categories not present in SemanticKITTI.

\begin{figure*}[t]
\begin{center}
\includegraphics[width=\linewidth]{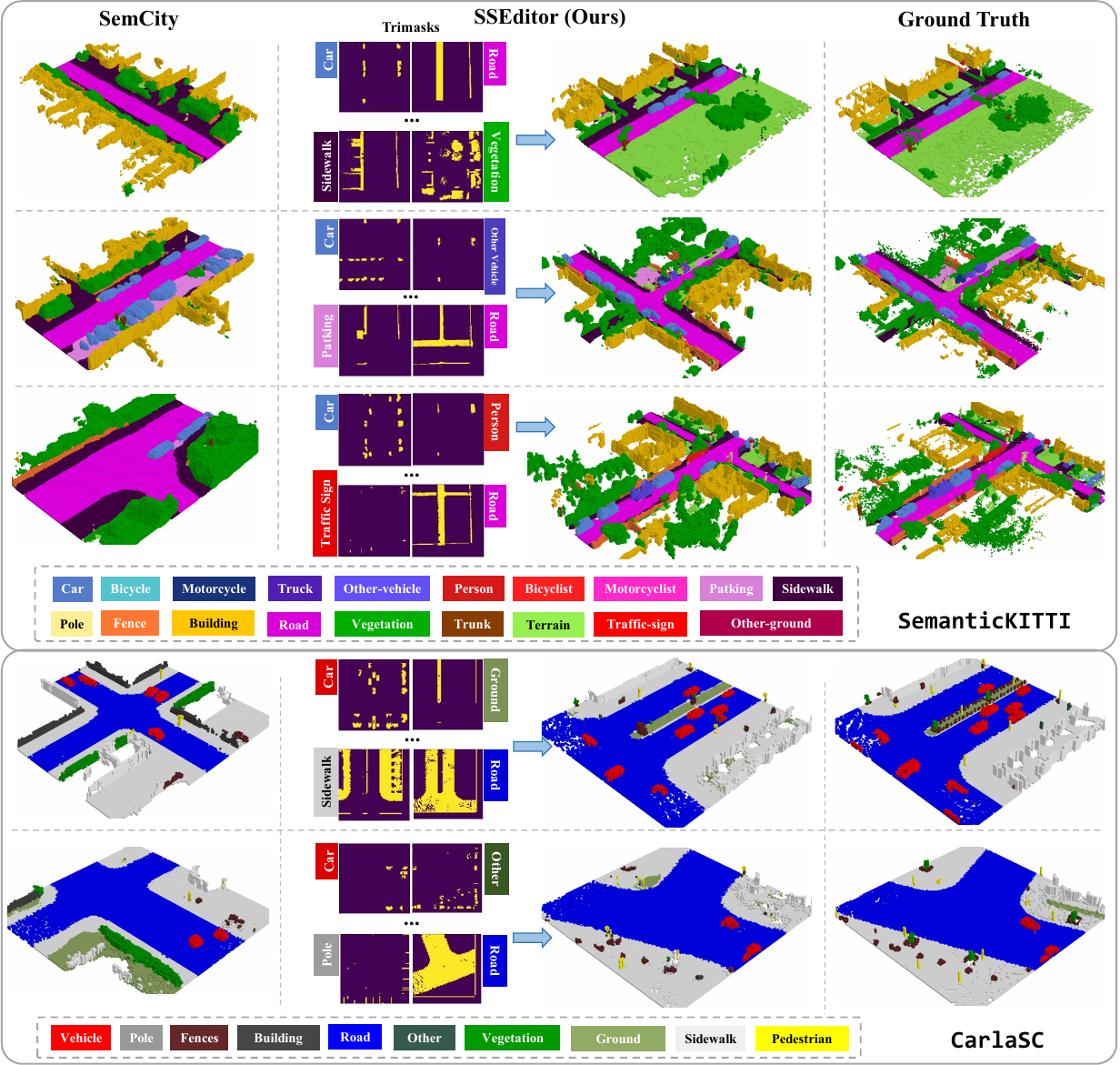}
\end{center}
\caption{Visualization of semantic scene generation comparing with SemCity \cite{lee2024semcity} on SemanticKITTI \cite{behley2019semantickitti} and CalarSC \cite{wilson2022motionsc}. Under the guidance of the trimask as a condition, SSEditor demonstrates its strong controllability.}
\label{fig:vis_generation}
\end{figure*}

\subsection{Implementation Details}
All experiments are conducted on a single NVIDIA RTX 3090-24G GPU. For the 3D scene autoencoder, the batch size is set to 4, while for the controllable mask-to-scene generation, the batch size is set to 1. The downsampling factors are configured as $d=2$ and $d_z=1$. The loss weight $\alpha$ in the Eq. \ref{eq1} is set to 1, the latent channel of triplane features $\mathcal{T}$ equals 16 and the embedding channel $C_{emb}=64$. The learning rate for the autoencoder is 1e-3, while the learning rate for the diffusion model is 1e-4. Following the settings of \cite{lee2024semcity, lee2023diffusion}, the sampling time steps is set to 100 during both training and testing of the diffusion model. We utilize DDPM sampling strategy \cite{ho2020denoising} for downstream tasks, omitting the need for the resampling strategy in RePaint \cite{lugmayr2022repaint}.

\subsection{Evaluation Metrics}
We adopt evaluation metrics from prior works \cite{tang2024diffuscene, zhai2024commonscenes,lee2024semcity} rendering 3D scenes into 2D images and use traditional 2D evaluation metrics to assess the quality and diversity of generated scenes:

\noindent\textbf{Fr\'{e}chet Inception Distance (FID)} \cite{heusel2017gans} measures the similarity between the real and generated data distributions by comparing their feature statistics in the latent space of the ImageNet-pretrained Inception network.

\noindent\textbf{Inception Score (IS)} \cite{salimans2016improved} evaluates both the quality and diversity of generated samples by computing a statistical
score from the Inception network.

\noindent\textbf{Kernel Inception Distance (KID)} \cite{binkowski2018demystifying} computes the squared Maximum Mean Discrepancy (MMD) between the real and generated data distributions using features extracted from the Inception network.

\noindent\textbf{Precision} measures the proportion of generated samples that fall within the support of the real data distribution, while \textbf{Recall} measures the proportion of the real data distribution covered by the generated samples.

In addition, we use the intersection over union (\textbf{IoU}) and mean IOU (\textbf{mIoU}) metrics to evaluate the overall scene reconstruction quality and the reconstruction quality for each class, respectively.

\begin{figure}[ht]
\begin{center}
\includegraphics[width=\linewidth]{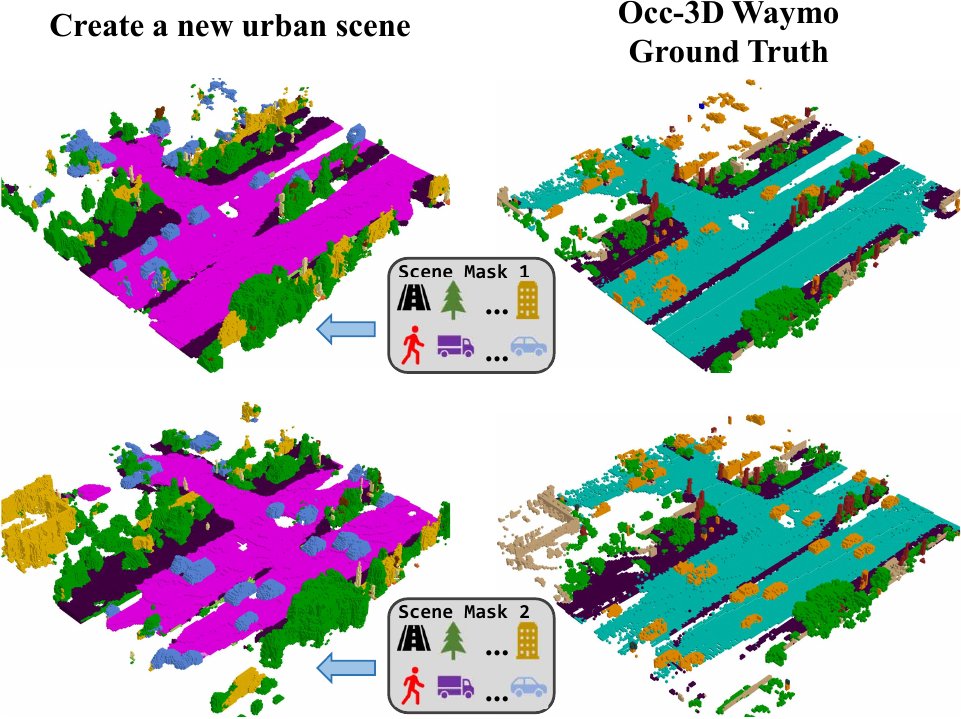}
\end{center}
\caption{Create a novel urban scene from masks. The novel scene generation is tested on the unseen Occ-3D Waymo dataset \cite{tian2024occ3d}.}
\label{fig:gen_new_scene}
\end{figure}

\subsection{Quantitative Results}

\textbf{Generation.} Table \ref{tab1} provides quantitative results on SmeanticKITTI and CarlaSC comparing with SSD \cite{lee2023diffusion} and SemCity \cite{lee2024semcity}. In overall generation quality and diversity, our SSEditor outperforms the previous methods \cite{lee2023diffusion, lee2024semcity} on SemanticKITTI \cite{behley2019semantickitti}, particularly in FID and recall, where we achieve improvements of 21.68\% and 39\%, respectively, compared to SemCity. On CarlaSC \cite{wilson2022motionsc}, SSEditor leads in all metrics except for IS, with FID improving by 63.04\% over SemCity. Note that SemCity do not disclose which image sets are used for evaluation, making the results non-reproducible. To ensure a fair comparison, we train on the training set and generate scenes on the validation set to obtain the evaluation results.

\noindent\textbf{Semantic Scene Completion.} We assess the controllability and scene reconstruction capabilities of our method through semantic scene completion. Table \ref{tab2} demonstrates that SSEditor performs well on the SemanticKITTI validation set. We only reference two state-of-the-art methods from different modalities, as other unconditional diffusion models \cite{lee2024semcity,lee2023diffusion} lack the ability to reconstruct 3D semantic scenes. The IoU metric indicates that our method provides strong control over the position and size of objects during scene generation, while the mIoU score reflects a robust understanding of the semantics of the generated objects.

\subsection{Qualitative Results}
\noindent\textbf{Generation.} Fig. \ref{fig:vis_generation} showcases the qualitative results of the proposed SSEditor and SemCity \cite{lee2024semcity} on the SemanticKITTI \cite{behley2019semantickitti} and CarlaSC \cite{wilson2022motionsc} datasets. While SemCity \cite{lee2024semcity} effectively generates a variety of scenes using triplane representations, it lacks sufficient control, making scene customization challenging. In contrast, SSEditor allows for precise generation of 3D scenes guided by masks, offering enhanced controllability. In Fig. \ref{fig:vis_generation}, we create trimasks based on ground truth to verify our method's controllability. The results demonstrate that SSEditor excels in controlling both the overall background (e.g., road, vegetation) and specific objects (e.g., vehicles, pedestrians).

\noindent\textbf{Scene Editing.} Fig. \ref{fig:edit_detail} highlights the details of scene editing with SSEditor. By setting the trimask of a target object or background to zero, we can effectively remove it from the scene. We can also edit background assets for more realistic scenarios, like creating four-lane or eight-lane assets. Once the background is adjusted, we can add objects, like increasing the number of cars to simulate higher traffic volumes, to create more dynamic scenarios.

\noindent\textbf{Novel Scene Generation.} To further validate the controllability of SSEditor in generating new scenes, we apply the trained model to the Occ-3D Waymo dataset \cite{tian2024occ3d}. We adjust the trimasks from Occ-3D Waymo through interpolation to align with the standard size of trimasks in our asset library, due to the different resolutions of the datasets. Note that we only create trimasks for categories that appear in SemanticKITTI \cite{behley2019semantickitti}. The generated results in Fig. \ref{fig:gen_new_scene} demonstrate that SSEditor can effectively adapt to new scene generation, enabling the rapid creation of urban environments.

\subsection{Ablation Studies}

We conduct ablation experiments on the SemanticKITTI \cite{behley2019semantickitti} validation set to assess the contribution of each component of SSEditor, as shown in Table \ref{tab3}.

First, we evaluate the effectiveness of the geometric branch by retaining the semantic branch and concatenating the trimask with the noised triplane $\mathcal{T}_t$ as input. Next, we remove the semantic branch, followed by the semantic tokens within the branch, to examine their individual impact. Finally, we input only the noised triplane $\mathcal{T}_t$ to assess the role of concatenating the trimask. In all ablation experiments, removing any component results in a performance drop, highlighting the necessity of each component for optimal performance.

Additionally, as shown in Table \ref{tab:time}, we compared two sampling strategies: DDPM \cite{ho2020denoising} and the resampling technique from RePaint \cite{lugmayr2022repaint}. While resampling improves object integration with the environment during generation, it greatly increases inference time for 3D scene generation. In contrast, our method employs traditional DDPM sampling, which maintains high quality and controllability in both scene inpainting and outpainting, while reducing inference time.

\begin{table}[t]
\centering
\resizebox{\linewidth}{!}{
\begin{tabular}{l|ccccc}
\toprule[1pt]
Method & FID $\downarrow$ & KID $\downarrow$ & IS $\uparrow$ & Prec. $\uparrow$ & Rec. $\uparrow$ \\ \midrule
w/o geometric branch  & 60.32 & 0.05 & 2.45 $\pm$ 0.15 & 0.24 & 0.28 \\
w/o semantic branch  & 54.96 & 0.05 & 2.49 $\pm$ 0.13 & 0.27 & 0.37 \\
w/o semantic tokens  & 53.67 & 0.04 & 2.49 $\pm$ 0.12 & 0.27 & 0.38 \\
w/o mask concat  & 54.08 & 0.04 & 2.43 $\pm$ 0.17 & 0.23 & 0.19 \\
SSEditor (ours)  & \textbf{47.93} & \textbf{0.03} & \textbf{2.55 $\pm$ 0.14} & \textbf{0.31} & \textbf{0.51} \\ 
\bottomrule[1pt]
\end{tabular}
}
\caption{Ablation studies on scene generation. We validated the effectiveness of the geometric branch, semantic branch, and the concatenated input of the trimask on SemanticKITTI \cite{behley2019semantickitti}.}
\label{tab3}
\end{table}

\begin{table}[t]
\centering
\resizebox{\linewidth}{!}{
\begin{tabular}{l|c|cc}
\toprule[1pt]
Method & Steps & Sampling  & Inference Time \\ \midrule
\multirow{6}{*}{SSEditor}  &\multirow{2}{*}{100}  & Resampling \cite{lugmayr2022repaint} & 56.44s \\
  & & DDPM \cite{ho2020denoising} & 13.40s  \\ \cline{2-4}
  
  &\multirow{2}{*}{20}  & Resampling \cite{lugmayr2022repaint} & 13.89s \\
  & & DDPM \cite{ho2020denoising} & 3.66s  \\ \cline{2-4}
  
  &\multirow{2}{*}{10}  & Resampling \cite{lugmayr2022repaint} & 6.91s \\
  & & DDPM \cite{ho2020denoising} & 2.08s  \\

\bottomrule[1pt]
\end{tabular}
}
\caption{Ablation studies on sampling strategy. The inference time is reported based on 100 sample runs.}
\label{tab:time}
\end{table}
\section{Limitations}
Although SSEditor demonstrates strong capabilities for controllable scene generation, it still faces challenges with generating small objects, such as bicyclists and pedestrians. The generated areas sometimes contain incorrectly classified voxels, and the model's performance is highly sensitive to surrounding objects, which can lead to inaccuracies.These issues negatively affect the performance of downstream tasks that rely on high-quality scene generation. Predicting small objects in semantic scene completion is inherently challenging due to their low visibility and the complex interactions they have with the environment, resulting in lower mIoU performance. Future work could focus on addressing the long-tail distribution of data by incorporating more robust methods for representing and detecting small objects, as well as developing more fine-grained representation techniques that can improve the handling of these challenging cases.

\section{Conclusion}
In this paper, we propose SSEditor, a two-stage controllable scene generation framework based on the diffusion model. In the first stage, we leverage a 3D scene autoencoder to learn triplane representations. We then create a trimask asset library as a preparatory step for the second phase of training. In the second stage, we train a mask-conditional diffusion model for mask-to-scene generation, incorporating a geometric-semantic fusion module to enhance the extraction of geometric and semantic information. Experimental results on SemanticKITTI, CarlaSC, and Occ-3D Waymo demonstrate that our method outperforms existing unconditional diffusion approaches, offering superior controllability and high-quality scene generation. Moreover, SSEditor supports a wide range of applications, including the generation of novel 3D urban scenes (such as cross-dataset generation and road widening), controllable generation of dynamic objects, and scene outpainting.

\clearpage
{
    \small
    \bibliographystyle{ieeenat_fullname}
    \bibliography{main}
}


\end{document}